\documentclass[10pt,twocolumn,letterpaper]{article}

\usepackage[pagenumbers]{cvpr} 
\usepackage{times}
\usepackage{epsfig}
\usepackage{graphicx}
\usepackage{amsmath}
\usepackage{amssymb}

\usepackage[dvipsnames]{xcolor}
\definecolor{olivegreen}{HTML}{3C8031}

\definecolor{cvprblue}{rgb}{0.21,0.49,0.74}
\usepackage[pagebackref,breaklinks,colorlinks,citecolor=cvprblue]{hyperref}

\usepackage{booktabs}
\usepackage{adjustbox}
\usepackage{algorithm}
\usepackage{listings}
\usepackage{multirow}

\begin{document}

\title{Rethinking Saliency-Guided Weakly-Supervised Semantic Segmentation}

\author{
Beomyoung Kim$^{1,2}$\hspace{1.5em}Donghyun Kim$^{1,2}$\hspace{1.5em}Sung Ju Hwang$^{2}$\\ \\
{NAVER Cloud$^1$\hspace{3em}KAIST$^2$}\\
}

\maketitle
\begin{abstract}

This paper presents a fresh perspective on the role of saliency maps in weakly-supervised semantic segmentation (WSSS) and offers new insights and research directions based on our empirical findings. 
We conduct comprehensive experiments and observe that the quality of the saliency map is a critical factor in saliency-guided WSSS approaches. Nonetheless, we find that the saliency maps used in previous works are often arbitrarily chosen, despite their significant impact on WSSS.
Additionally, we observe that the choice of the threshold, which has received less attention before, is non-trivial in WSSS.
To facilitate more meaningful and rigorous research for saliency-guided WSSS, we introduce \texttt{WSSS-BED}, a standardized framework for conducting research under unified conditions. \texttt{WSSS-BED} provides various saliency maps and activation maps for seven WSSS methods, as well as saliency maps from unsupervised salient object detection models.
\end{abstract}

\section{Introduction}

Semantic segmentation is a fundamental computer vision task, and recent deep-learning based methods~\cite{(Mask2Former)cheng2022masked,(MaskFormer)cheng2021per,(HRViT)gu2022multi,(MaskDINO)li2022mask,(swinv2)liu2022swin,(ConvNext)liu2022convnet,(Segmenter)strudel2021segmenter,(HRNet)wang2020deep,(UPerNet)xiao2018unified,(SegFormer)xie2021segformer,(OCRNet)yuan2020object,(K-Net)zhang2021k} have demonstrated remarkable performance. Nonetheless, training semantic segmentation models requires delicate annotations at the pixel level, which can be a labor-intensive process. 
In an effort to reduce the annotation cost, weakly-supervised semantic segmentation (WSSS) has recently become a topic of significant interest.
While a variety of weak supervision sources such as bounding boxes~\cite{khoreva2017simple,lee2021bbam}, points~\cite{bearman2016s,kim2022beyond}, and scribbles~\cite{lin2016scribblesup,pan2021scribble} have been explored, WSSS utilizing image-level labels has gained a lot of attention due to its cost-effectiveness. In this paper, we solely focus on image-level supervised semantic segmentation.

Most existing WSSS approaches utilize class-wise activation maps, CAM~\cite{(CAM)zhou2016learning}, which are generated by the classifier trained with image-level labels only.
The activation maps are employed to generate pseudo segmentation labels that are used to train the segmentation network ($e.g.,$ DeepLab-V2~\cite{(DLV2)chen2017deeplab}).
The most important step in WSSS is generating pseudo labels using the activation maps.

\begin{figure}[t]
    \centering
    \includegraphics[width=\linewidth]{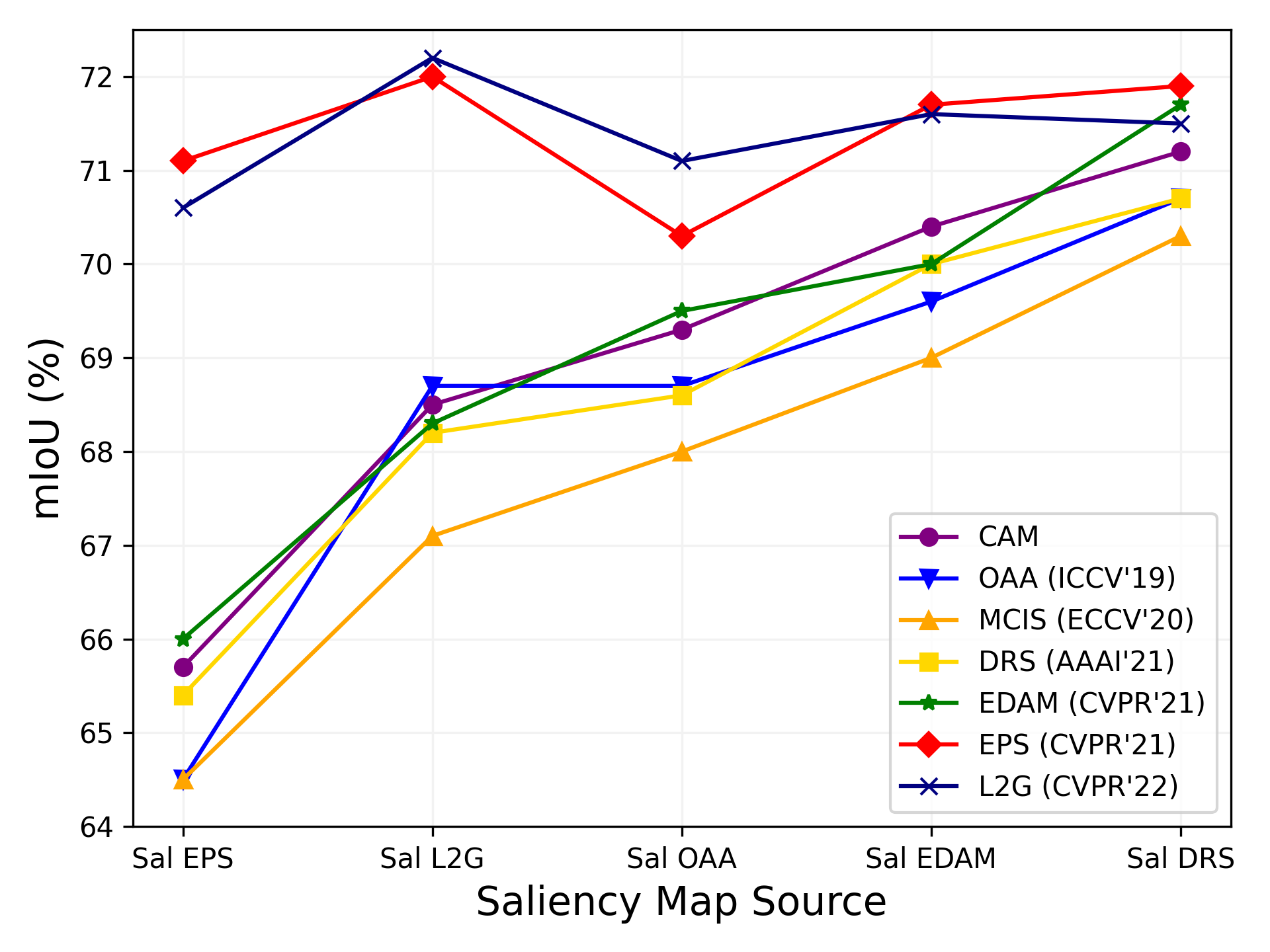}
    \caption{\textbf{Performance variation according to the saliency map.} We reveal the saliency map used in each method is not unified, and the impact of the saliency map on each method is highly significant. `Sal \texttt{METHOD}' on the x-axis denotes saliency maps used in \texttt{METHOD}. The scores are measured on VOC 2012 validation set.}
    \label{fig:cross_validation}
\end{figure}

\begin{figure*}[t]
    \centering
    \includegraphics[width=0.95\linewidth]{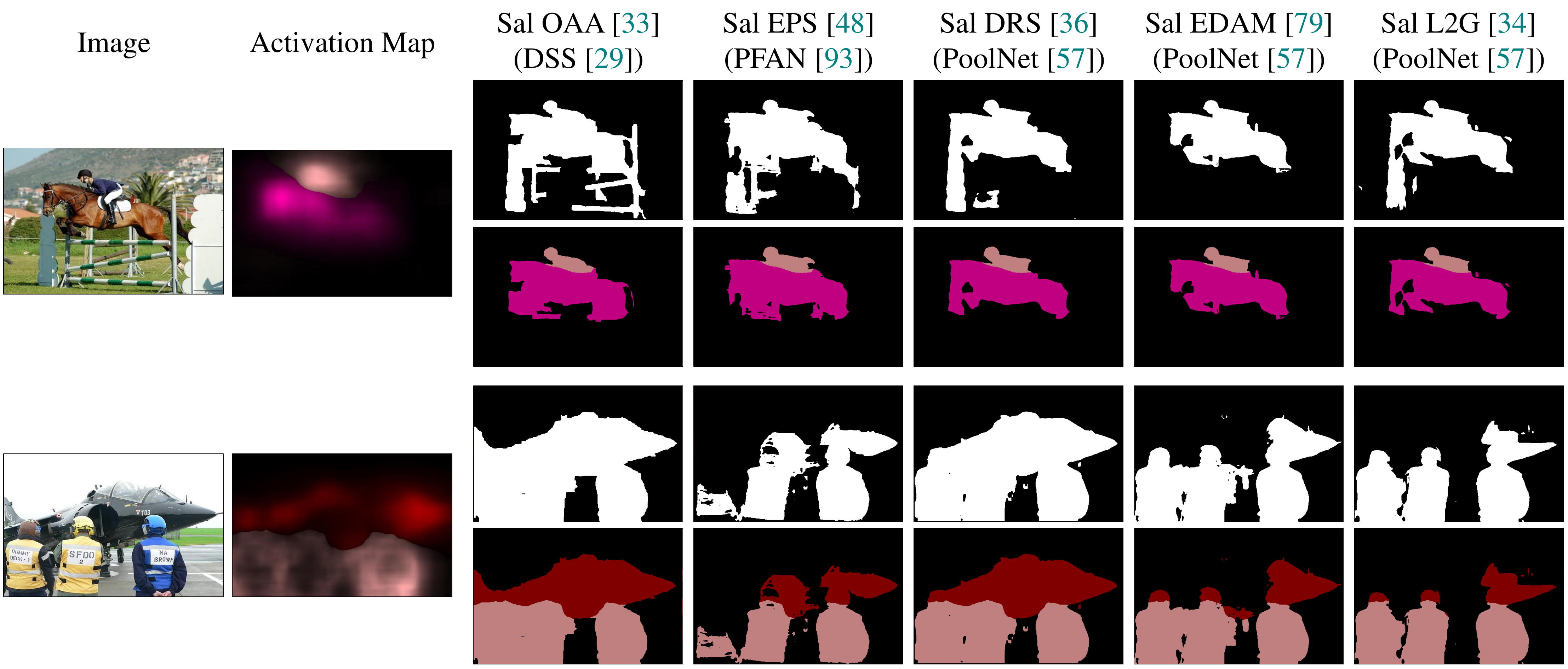}
     \caption{\textbf{Qualitative samples of pseudo segmentation labels (even rows) given saliency maps (odd rows) and activation maps.} `Sal \texttt{METHOD} (\texttt{SOD-MODEL})' indicates the saliency map used in WSSS \texttt{METHOD} and the pre-trained \texttt{SOD-MODEL} is employed when generating the saliency map. The saliency map is crucial in determining the quality of the pseudo label. Although some methods ($e.g.,$ DRS~\cite{(DRS)kim2021discriminative}, EDAM~\cite{(EDAM)wu2021embedded}, L2G~\cite{(L2G)jiang2022l2g}) employed the same \texttt{SOD-MODEL} ($i.e.,$ PoolNet~\cite{(PoolNet)liu2019simple}), the quality of the saliency map is different greatly.}
    \label{fig:sal_comparison}
\end{figure*}

The activation maps generated by CAM can identify discriminative object regions for each class.
However, since the classifier is not trained to recognize the background class, it often introduces substantial noise into the background region of the activation maps, leading to numerous false-positives in the generated pseudo labels.
Additionally, the activation maps tend to focus on sparse object regions, which can yield many false-negatives.

To address these issues, there have been several attempts~\cite{(AE-PSL)wei2017object,(DSRG)huang2018weakly,(FickleNet)lee2019ficklenet,(ICD)fan2020learning,(AuxSegNet)xu2021leveraging,(RCA)zhou2022regional} to distinguish between the foreground and background regions utilizing off-the-shelf saliency maps, which are referred to as a saliency-guided approach; we note that we solely cover the saliency-guided approaches in this paper.
In practice, the salient object detector has been employed in various tasks such as visual tracking~\cite{hong2015online} and image retrieval~\cite{gao20123} due to its free-accessibility and effectiveness as a general-purpose object mask provider.
By utilizing the saliency map generated from a salient object detector, they concentrate on activations in the salient region while filtering out noisy activations in the background.

\begin{figure*}[t]
    \centering
    \includegraphics[width=\linewidth]{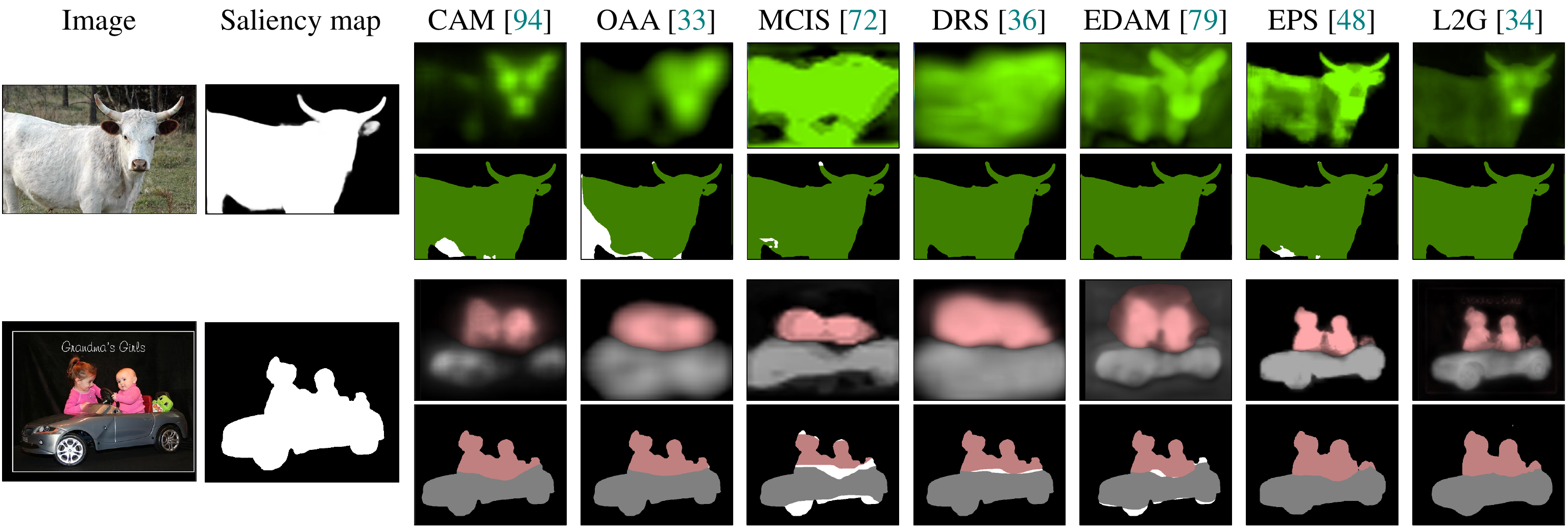}
    \caption{\textbf{Qualitative comparisons for activation maps (odd rows) and pseudo labels (even rows) of WSSS methods using the same saliency map.} The activation map from CAM~\cite{(CAM)zhou2016learning} appears to highlight the object region sparsely compared to other methods, but the quality of the pseudo label from CAM is highly competitive with other methods when using a lower threshold. Since the quality of the activation map of each method varies largely, the threshold is required to be differently set for each method.}
    \label{fig:cam_and_pseudo_labels}
\end{figure*}

In this paper, we present a fresh perspective on the role of saliency maps in WSSS and provide new insights, as well as future research directions for saliency-guided WSSS based on empirical observations. 
Based on comprehensive experiments, we observe the following:

\begin{enumerate}
    \item 
    Our observations suggest that \textbf{the quality of the saliency map is a crucial factor in saliency-guided WSSS approaches}. Specifically, we find that the choice of the pre-trained salient object detector (SOD) for generating the saliency maps as well as the dataset it is trained on, has a significant impact on the quality of saliency maps, and in turn, the performance of saliency-guided WSSS. Through comprehensive experiments using four SOD models and three SOD datasets, we show that high-quality saliency maps lead to boosted WSSS performance, even surpassing the state-of-the-art results. Conversely, low-quality saliency maps lead to a substantial degeneration in the performance of WSSS methods.
    \item
    Despite the significant impact of the saliency maps on its WSSS performance, \textbf{the saliency maps used in previous works are inconsistent.} For example, some works~\cite{(ICD)fan2020learning,(EME)fan2020employing,(SeeNet)hou2018self,(OAA)jiang2019integral,(groupWSSS)li2021group,(AuxSegNet)xu2021leveraging} use DSS~\cite{(DSS)hou2017deeply}, some~\cite{(L2G)jiang2022l2g,(DRS)kim2021discriminative,(MCIS)sun2020mining,(EDAM)wu2021embedded,(MOR)zhang2020splitting} use PoolNet~\cite{(PoolNet)liu2019simple}, and some~\cite{(PPC)du2022weakly,(EPS)lee2021railroad,(RCA)zhou2022regional} use PFAN~\cite{(PFAN)zhao2019pyramid} as their salient object detection (SOD) model. 
    Furthermore, most studies lack specifications of which SOD dataset was used for pre-training the SOD model. Even when they use the same SOD model, the quality of the saliency map varies across the methods (Figure~\ref{fig:sal_comparison}). 
    Our experiments using seven WSSS methods reveal that their performance varies widely when using different saliency maps (Figure~\ref{fig:cross_validation}).
    \item
    \textbf{The conventional CAM, often considered as a baseline, can achieve comparable performance to existing WSSS methods by using a lower threshold} (Figure~\ref{fig:threshold_graph}). Previous studies have paid little attention to the choice of threshold, and the threshold used previously was not optimal for the conventional CAM. A lower threshold enables the conventional CAM to capture class-wise information in the salient regions while filtering out noise in the background using the saliency map (Figure~\ref{fig:cam_and_pseudo_labels}).
\end{enumerate}

The goal of this paper is to encourage researchers to reconsider the role of the saliency map on WSSS and promote more meaningful and rigorous research for WSSS by presenting our empirical findings concerning the saliency map.
To facilitate this effort, we introduce a new framework called \textbf{\texttt{WSSS-BED}}\footnote{available at \url{https://github.com/clovaai/WSSS-BED}} that establishes a baseline framework for conducting research under unified conditions.
When a user submits their activation maps to \texttt{WSSS-BED}, it generates pseudo labels using our unified saliency maps and a specified threshold. Our \texttt{WSSS-BED} contains various saliency maps and activation maps for seven WSSS methods. Additionally, \texttt{WSSS-BED} provides saliency maps from unsupervised SOD models to facilitate the fully image-level supervised semantic segmentation without the use of any ground-truth object masks for pre-training the SOD model.
We are inspired by DomainBed~\cite{(domain-bed)gulrajani2021in}, and \texttt{WSSS-BED} will be continuously updated the available models upon pull request from fellow researchers.

\section{Preliminary}

This section provides a brief overview of the fundamental information required for saliency-guided weakly-supervised semantic segmentation (WSSS) approaches, including: (1) obtaining class activation maps, (2) obtaining saliency maps, (3) generating pseudo segmentation labels using saliency maps, and (4) training the baseline segmentation network using the pseudo labels.

\subsection{Notations}
For the saliency-guided WSSS, three kinds of networks are required: classification network $f_{cls}$ ($e.g.,$ ResNet-38~\cite{(wresnet)wu2019wider}), pre-trained salient object detection network $f_{sod}$ ($e.g.,$ PoolNet~\cite{(PoolNet)liu2019simple}), and segmentation network $f_{seg}$ ($e.g.,$ DeepLabV2~\cite{(DLV2)chen2017deeplab}).
In addition, two types of datasets are required: WSSS dataset $\mathcal{D}_{wseg}$ consisting of a set of pairs $(I^{n}, y^{n},S^{n})$ where $I^{n}$ is an input image, $y^{n}$ is a ground-truth image-level label, and $S^{n}$ is a saliency map; salient object detection dataset $\mathcal{D}_{sod}$ consisting of a set of pairs input images and ground-truth salient object masks. 

\subsection{Class Activation Map}
Most existing WSSS methods utilize the CAM~\cite{(CAM)zhou2016learning} to obtain class-wise activation maps.
Given the classification network $f_{cls}$, the activation map $A$ is directly computed from the feature maps of the last convolutional layer, $F \in \mathbb{R}^{C{\times}H{\times}W}$, where $C$ is the number of classes, $H$ and $W$ is the spatial height and width of the last feature maps:
\begin{equation}
    A_{c} = \frac{ReLU(F_{c})}{max(ReLU(F_c))}.
\end{equation}
The classification network is trained in a multi-label classification manner.
Namely, the logit of the network, $z \in \mathbb{R}^{C{\times}1}$, is obtained from the last feature maps $F$ followed by a global average pooling (GAP) layer.
Then, the loss function is defined as follows:
\begin{equation}
    L_{bce} = -\frac{1}{C} \sum_{c=1}^{C}{y_{c}\text{log}(\sigma(z_{c}))+(1-y_{c})\text{log}(1-\sigma(z_{c}))},
\end{equation}
where $\sigma$ is a sigmoid function, and $y$ is the one-hot ground-truth image-level label.

However, the activation map from the CAM appears to highlight only the sparse object region.
Therefore, most existing WSSS methods~\cite{(OAA)jiang2019integral,(L2G)jiang2022l2g,(DRS)kim2021discriminative,(EPS)lee2021railroad,(MCIS)sun2020mining,(EDAM)wu2021embedded} have tried to improve the activation map to cover the object region entirely, as shown in Figure~\ref{fig:cam_and_pseudo_labels}.

\subsection{Saliency Map Generation}
The salient object detection model $f_{sod}$ is often trained in a fully-supervised manner using the salient object detection dataset $\mathcal{D}_{sod}$.
Then, given the image in the WSSS dataset $I^{n}$, the pre-trained salient object detection model $f_{sod}$ generates the saliency map $S^{n}$: $S^{n}{=}f_{sod}(I^{n})$.

\begin{figure}[t]
    \centering
    \includegraphics[width=\linewidth]{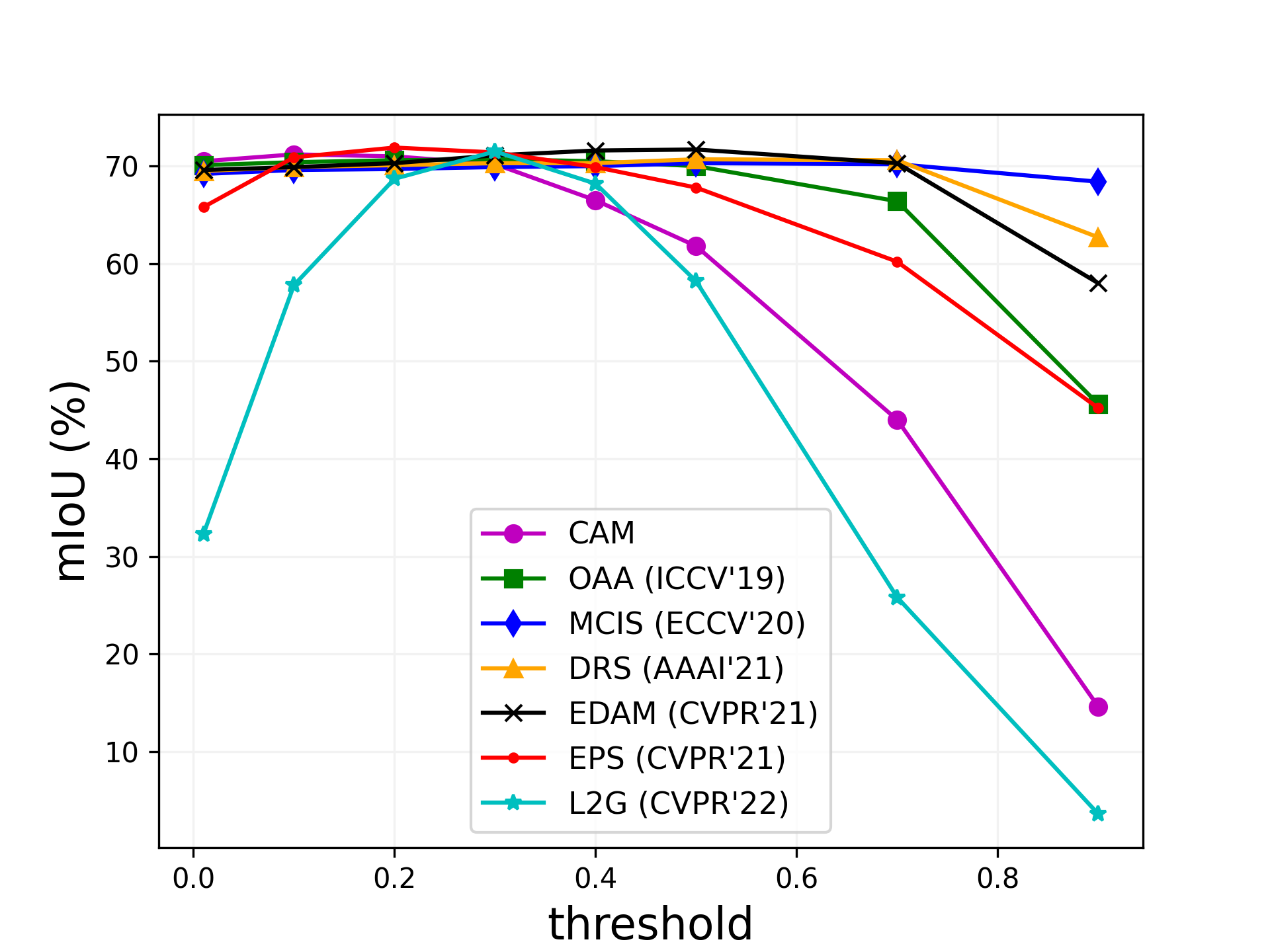}
    \caption{\textbf{Quantitative comparisons of WSSS methods according to the threshold $\tau$ given equivalent saliency maps}. The impact of the threshold on WSSS methods is highly substantial, and the optimal threshold varies from each method. The conventional CAM with a low threshold shows a highly competitive performance compared with state-of-the-art methods.}
    \label{fig:threshold_graph}
\end{figure}

\subsection{Pseudo Segmentation Label Generation}
The pseudo segmentation label is generated using the attention map $A^{n}$ and the saliency map $S^{n}$.
Specifically, pixels in $A^{n}$ that are larger than a threshold $\tau$ are regarded as object cues.
Also, given the saliency map, non-salient pixels in $A^{n}$ are regarded as background cues.
Pixels that are neither the object nor the background cues are ignored when training the segmentation network.

\begin{algorithm}[h]
   \caption{Python pseudo label generation pseudocode.}
    \definecolor{codeblue}{rgb}{0.25,0.5,0.5}
    \lstset{
      basicstyle=\fontsize{8pt}{8pt}\ttfamily\bfseries,
      commentstyle=\fontsize{8pt}{8pt}\color{codeblue},
      keywordstyle=\fontsize{8pt}{8pt},
    }
    \begin{lstlisting}[language=python]
    # A: attention map (CxHxW) [0, 1] 
    # S: saliency map (1xHxW) {0, 1}
    # P: pseudo label (HxW)
    # y: image-level label (Cx1x1) {0, 1}
    # tau: threshold
    
    bg = (1 - S) # backgorund cue
    fg = np.where(A > tau, A, 0) # object cue
    fg = fg * y # activate only valid classes

    P = np.concatenate([bg, fg], axis=0) #(C+1,H,W)
    P = np.argmax(P, axis=0) # (H,W)
    P[(S==1) & (P==0)] = 255 # ignore label
    \end{lstlisting}
    \label{alg:pseudo_code}
\end{algorithm}

\subsection{Training Segmentation Network}
Following the same training recipe for the fully-supervised semantic segmentation network, the network $f_{seg}$ is trained using the pseudo segmentation labels.
The WSSS methods are then evaluated by comparing the performance of $f_{seg}$ trained using their pseudo labels.
\begin{table*}[t]
  \begin{center}
  \begin{adjustbox}{max width=\linewidth}
  \begin{tabular}{c|c|c|c|c|c|c|c|c}
    \toprule
    SOD Model & SOD Dataset & CAM~\cite{(CAM)zhou2016learning}& OAA~\cite{(OAA)jiang2019integral} & MCIS~\cite{(MCIS)sun2020mining} & DRS~\cite{(DRS)kim2021discriminative} & EDAM~\cite{(EDAM)wu2021embedded} & EPS~\cite{(EPS)lee2021railroad} & L2G~\cite{(L2G)jiang2022l2g} \\
    \midrule
    \multicolumn{2}{c|}{\textit{Paper}} & - & 65.2 & 66.2 & 70.4 & 70.9 & 70.9 & 72.1 \\
    \midrule
     & MSRA-B~\cite{(MSRA)liu2010learning} & 63.2$\pm$0.3 & 62.2$\pm$0.1 & 61.8$\pm$0.4 & 63.2$\pm$0.4 & 63.2$\pm$0.3 & 70.1$\pm$0.1 & 70.8$\pm$0.0 \\
     & HKU-IS~\cite{(HKU)li2015visual}+MSRA-B & 69.7$\pm$0.1 & 68.9$\pm$0.2 & 68.1$\pm$0.1 & 69.1$\pm$0.1 & 69.9$\pm$0.1 & 71.4$\pm$0.2 & 71.1$\pm$0.1 \\
    DSS~\cite{(DSS)hou2017deeply} & DUTS~\cite{(DUTS)wang2017learning} & 67.8$\pm$0.1 & 67.0$\pm$0.1 & 66.4$\pm$0.1 & 67.3$\pm$0.2 & 67.5$\pm$0.2 & 71.2$\pm$0.2 & 71.4$\pm$0.2 \\
    \textsubscript{(CVPR'17)} & COCO~\cite{(coco)lin2014microsoft} & 73.1$\pm$0.1 & 72.6$\pm$0.3 & 71.6$\pm$0.3 & 71.9$\pm$0.1 & 73.4$\pm$0.2 & 72.5$\pm$0.1 & 71.7$\pm$0.1 \\
     & COCO-20~\cite{(coco)lin2014microsoft} & 72.1$\pm$0.1 & 72.1$\pm$0.3 & 71.1$\pm$0.3 & 71.3$\pm$0.1 & 72.4$\pm$0.1 & 73.9$\pm$0.2 & 72.4$\pm$0.0 \\
    \midrule
     & MSRA-B~\cite{(MSRA)liu2010learning} & 62.1$\pm$0.7 & 60.3$\pm$0.4 & 60.0$\pm$0.4 & 61.7$\pm$0.6 & 61.6$\pm$0.5 & 71.4$\pm$0.2 & 70.9$\pm$0.2 \\
     & HKU-IS~\cite{(HKU)li2015visual}+MSRA-B & 67.6$\pm$0.1 & 66.5$\pm$0.1 & 65.6$\pm$0.3 & 67.4$\pm$0.2 & 67.4$\pm$0.1 & 71.9$\pm$0.1 & 71.3$\pm$0.2 \\
    PFAN~\cite{(PFAN)zhao2019pyramid} & DUTS~\cite{(DUTS)wang2017learning} & 68.0$\pm$0.3 & 66.9$\pm$0.2 & 66.4$\pm$0.0 & 67.4$\pm$0.2 & 67.4$\pm$0.1 & 72.1$\pm$0.1 & 71.6$\pm$0.3 \\
    \textsubscript{(CVPR'19)} & COCO~\cite{(coco)lin2014microsoft} & 72.0$\pm$0.2 & 71.7$\pm$0.2 & 70.8$\pm$0.1 & 70.8$\pm$0.4 & 72.3$\pm$0.3 & 71.6$\pm$0.1 & 71.5$\pm$0.3 \\
     & COCO-20~\cite{(coco)lin2014microsoft} & 72.1$\pm$0.2 & 72.1$\pm$0.4 & 71.1$\pm$0.2 & 71.6$\pm$0.4 & 72.0$\pm$0.2 & 73.7$\pm$0.2 & 72.3$\pm$0.1 \\
    \midrule
     & MSRA-B~\cite{(MSRA)liu2010learning} & 62.9$\pm$0.6 & 61.4$\pm$0.4 & 61.1$\pm$0.7 & 62.7$\pm$0.7 & 62.4$\pm$0.8 & 71.4$\pm$0.1 & 71.3$\pm$0.3 \\
     & HKU-IS~\cite{(HKU)li2015visual}+MSRA-B & 69.5$\pm$0.3 & 68.8$\pm$0.1 & 68.1$\pm$0.1 & 69.0$\pm$0.3 & 69.5$\pm$0.2 & 71.9$\pm$0.1 & 71.3$\pm$0.3 \\
    PoolNet~\cite{(PoolNet)liu2019simple} & DUTS~\cite{(DUTS)wang2017learning} & 68.5$\pm$0.1 & 67.0$\pm$0.3 & 67.2$\pm$0.3 & 68.1$\pm$0.2 & 68.5$\pm$0.3 &  72.7$\pm$0.2 & 71.7$\pm$0.2 \\
    \textsubscript{(CVPR'19)} & COCO~\cite{(coco)lin2014microsoft} & 73.9$\pm$0.1 & 73.1$\pm$0.1 & 71.8$\pm$0.4 & 72.4$\pm$0.3 & 74.0$\pm$0.2 & 72.4$\pm$0.0 & 71.7$\pm$0.2 \\
     & COCO-20~\cite{(coco)lin2014microsoft} & 74.4$\pm$0.1 & 74.0$\pm$0.1 & 72.6$\pm$0.2 & 72.5$\pm$0.2 & 74.3$\pm$0.2 & 74.2$\pm$0.0 & 72.8$\pm$0.1 \\
    \midrule
     & MSRA-B~\cite{(MSRA)liu2010learning} & 67.4$\pm$0.3 & 66.4$\pm$0.3 & 66.0$\pm$0.2 & 67.2$\pm$0.0 & 67.8$\pm$0.1 & 71.5$\pm$0.1 & 71.1$\pm$0.2 \\
     & HKU-IS~\cite{(HKU)li2015visual}+MSRA-B & 70.1$\pm$0.3 & 69.4$\pm$0.4 & 68.8$\pm$0.2 & 70.0$\pm$0.3 & 70.3$\pm$0.4 & 72.2$\pm$0.2 & 71.3$\pm$0.2 \\
    VST~\cite{(VST)liu2021visual} & DUTS~\cite{(DUTS)wang2017learning} & 70.5$\pm$0.2 & 69.6$\pm$0.2 & 69.2$\pm$0.2 & 70.0$\pm$0.1 & 70.3$\pm$0.1 & 72.4$\pm$0.2 & 71.9$\pm$0.2 \\
    \textsubscript{(ICCV'21)} & COCO~\cite{(coco)lin2014microsoft} & 73.9$\pm$0.2 & 74.3$\pm$0.1 & 73.1$\pm$0.1 & 73.1$\pm$0.2 & 74.8$\pm$0.2 & 72.8$\pm$0.1 & 71.5$\pm$0.1 \\
     & COCO-20~\cite{(coco)lin2014microsoft} & 74.5$\pm$0.2 & 74.6$\pm$0.1 & 73.5$\pm$0.2 & 73.8$\pm$0.1 & \textbf{74.9$\pm$0.1} & 74.3$\pm$0.1 & 73.0$\pm$0.0 \\
     \midrule
    DeepUSPS~\cite{(DeepUSPS)nguyen2019deepusps} & - (\textit{unsupervised}) & 61.6$\pm$0.4 & 60.6$\pm$0.3 & 60.3$\pm$0.2 & 61.3$\pm$0.2 & 62.0$\pm$0.4 & 69.3$\pm$0.2 & 70.4$\pm$0.2 \\
    MOVE~\cite{(MOVE)bielski2022move} & - (\textit{unsupervised}) & 69.3$\pm$0.1 & 68.2$\pm$0.1 & 67.9$\pm$0.0 & 68.9$\pm$0.3 & 69.3$\pm$0.3 & 71.0$\pm$0.1 & 71.0$\pm$0.2 \\     
    \bottomrule
  \end{tabular}
  \end{adjustbox}
  \end{center}
  \vspace{-5mm}
  \caption{
    \textbf{Quantitative results of WSSS methods on VOC 2012 validation set according to the saliency map generated using different SOD models and SOD datasets.} The \textit{Paper} in the 2nd row means the score is from their paper. We re-evaluate the performance of seven WSSS methods using unified saliency maps. The result varies significantly depending on the SOD model and dataset. Note that the results are measured using DeepLabV2~\cite{(DLV2)chen2017deeplab} with ResNet-101~\cite{(resnet)he2016deep} backbone whose fully-supervised performance on VOC 2012 is 77.9\%.
  }
  \label{tab:result_voc2012}
  \vspace{-2mm}
\end{table*}

\section{Experiments}

\subsection{Datasets}
\paragraph{\textbf{Segmentation Datasets.}}
We use two popular semantic segmentation datasets, PASCAL VOC 2012~\cite{(voc2012)everingham2015pascal} and MS COCO 2014~\cite{(coco)lin2014microsoft}.
For VOC 2012, we leverage an augmented version~\cite{hariharan2011semantic} consisting of 10,582 training and 1449 validation images with 20 classes. 
COCO 2014 dataset consists of 82,081 training and 40,137 validation images across 80 classes. 
We use only image-level labels for both datasets.

\begin{figure}[t]
    \centering
    \includegraphics[width=\linewidth]{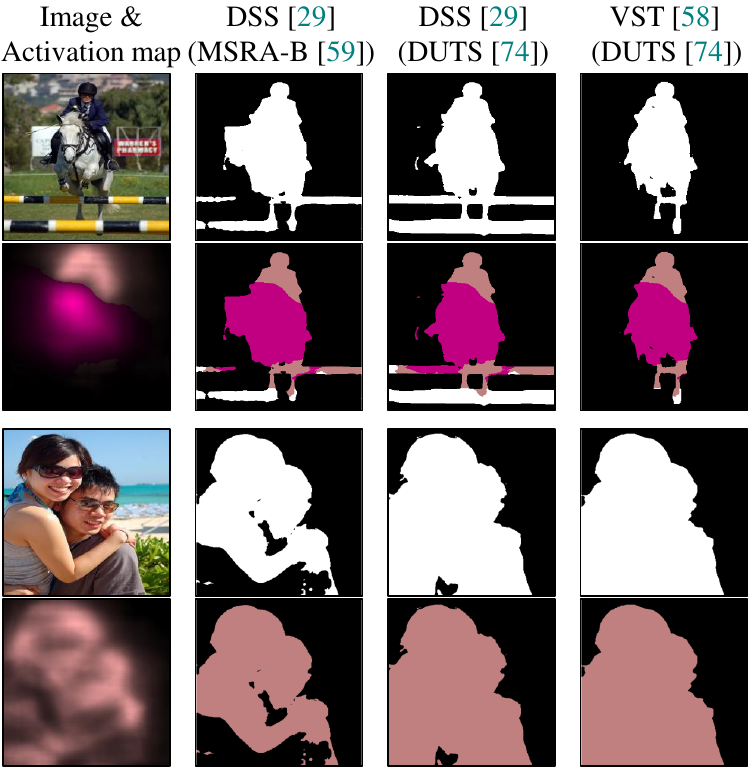}
    \caption{\textbf{Qualitative samples for saliency maps and pseudo labels according to the SOD model and dataset.} `\texttt{SOD-MODEL} (\texttt{SOD-DATASET})' denotes the saliency map is generated from \texttt{SOD-MODEL} that is pre-trained on \texttt{SOD-DATASET}. When using the larger-scale dataset ($e.g.,$ DUTS~\cite{(DUTS)wang2017learning}) or the more powerful model ($e.g.,$ VST~\cite{(VST)liu2021visual}), the quality of the generated saliency map improves accordingly, resulting in the higher-quality pseudo label.}
    \label{fig:pseudo_label_sod_model_dataset}
\end{figure}

\paragraph{\textbf{Salient Object Detection Datasets.}}
We employ three commonly used salient object detection (SOD) datasets, namely MSRA-B~\cite{(MSRA)liu2010learning}, HKU-IS~\cite{(HKU)li2015visual}, DUTS~\cite{(DUTS)wang2017learning}.
The MSRA-B, HKU-IS, and DUTS datasets consist of 5000, 4447, and 15,572 images, respectively. 
We combine HKU-IS and MSRA-B datasets to create a dataset consisting of 9447 images. 
Each dataset comprises pairs of images and class-agnostic salient object masks.

\subsection{Implementation Details}
\paragraph{\textbf{WSSS Methods.}}
We adopt seven baseline methods, namely CAM~\cite{(CAM)zhou2016learning}, OAA~\cite{(OAA)jiang2019integral}, MCIS~\cite{(MCIS)sun2020mining}, DRS~\cite{(DRS)kim2021discriminative}, EDAM~\cite{(EDAM)wu2021embedded}, EPS~\cite{(EPS)lee2021railroad}, and L2G~\cite{(L2G)jiang2022l2g}.
For each WSSS approach, we strictly follow their official implementation details.
We collect activation maps of each method and generate pseudo labels as described in Algorithm~\ref{alg:pseudo_code}.
We note that EPS~\cite{(EPS)lee2021railroad} and L2G~\cite{(L2G)jiang2022l2g} do not use saliency maps when generating pseudo labels; they use only thresholding on the activation map and we discuss it in Section~\ref{sec:discussion}.

\paragraph{\textbf{SOD Models.}}
For SOD models, we adopt commonly used SOD models in WSSS research fields, namely DSS~\cite{(DSS)hou2017deeply}, PFAN~\cite{(PFAN)zhao2019pyramid}, and PoolNet~\cite{(PoolNet)liu2019simple}.
We additionally employ the recent SOD model, VST~\cite{(VST)liu2021visual}, to further supplement the effect of SOD models on WSSS. 
We train four SOD models for each SOD dataset, strictly following their official implementation details. 
In addition, we utilize two unsupervised SOD models, DeepUSPS~\cite{(DeepUSPS)nguyen2019deepusps} and MOVE~\cite{(MOVE)bielski2022move}, which do not require ground-truth masks for supervision.

\begin{table*}[t]
  \begin{center}
  \begin{adjustbox}{max width=0.75\linewidth}
  \begin{tabular}{c|c|c|c|c|c}
    \toprule
    SOD Model & SOD Dataset & CAM~\cite{(CAM)zhou2016learning} & DRS~\cite{(DRS)kim2021discriminative} & EPS~\cite{(EPS)lee2021railroad} & L2G~\cite{(L2G)jiang2022l2g} \\
    \midrule
    \multicolumn{2}{c|}{\textit{Paper}} & - & - & 35.7$^{\dagger}$ & 44.2 \\
    \midrule
     & MSRA-B~\cite{(MSRA)liu2010learning} & 35.0$\pm$0.4 & 31.8$\pm$0.3 & 37.0$\pm$0.2 & 40.8$\pm$0.3 \\
    DSS~\cite{(DSS)hou2017deeply} & HKU-IS~\cite{(HKU)li2015visual}+MSRA-B & 41.0$\pm$0.1 & 37.5$\pm$0.2 & 38.9$\pm$0.1 & 43.3$\pm$0.1 \\
    \textsubscript{(CVPR'17)} & DUTS~\cite{(DUTS)wang2017learning} & 39.6$\pm$0.2 & 36.3$\pm$0.3 & 40.4$\pm$0.0 & 43.6$\pm$0.2 \\
    \midrule
     & MSRA-B~\cite{(MSRA)liu2010learning} & 33.0$\pm$0.4 & 29.7$\pm$0.4 & 39.3$\pm$0.1 & 42.7$\pm$0.1\\
    PFAN~\cite{(PFAN)zhao2019pyramid} & HKU-IS~\cite{(HKU)li2015visual}+MSRA-B & 38.3$\pm$0.3 & 34.9$\pm$0.2 & 39.8$\pm$0.1 & 41.6$\pm$0.1 \\
    \textsubscript{(CVPR'19)} & DUTS~\cite{(DUTS)wang2017learning} & 39.3$\pm$0.3 & 36.3$\pm$0.2 & 41.4$\pm$0.0 & 44.2$\pm$0.1 \\
    \midrule
     & MSRA-B~\cite{(MSRA)liu2010learning} & 31.7$\pm$0.5 & 28.4$\pm$1.0 & 40.3$\pm$0.1 & 41.8$\pm$0.1 \\
    PoolNet~\cite{(PoolNet)liu2019simple} & HKU-IS~\cite{(HKU)li2015visual}+MSRA-B & 39.9$\pm$0.4 & 36.6$\pm$0.1 & 41.8$\pm$0.0 & 43.9$\pm$0.0 \\
    \textsubscript{(CVPR'19)} & DUTS~\cite{(DUTS)wang2017learning} & 38.7$\pm$0.3 & 35.7$\pm$0.3 & 43.2$\pm$0.0 & 43.9$\pm$0.1 \\
    \midrule
     & MSRA-B~\cite{(MSRA)liu2010learning} & 39.8$\pm$0.4 & 36.6$\pm$0.2 & 41.3$\pm$0.0 & 44.5$\pm$0.1 \\
    VST~\cite{(VST)liu2021visual} & HKU-IS~\cite{(HKU)li2015visual}+MSRA-B & 42.0$\pm$0.2 & 38.9$\pm$0.2 & 42.5$\pm$0.0 & 44.8$\pm$0.0 \\
    \textsubscript{(ICCV'21)} & DUTS~\cite{(DUTS)wang2017learning} & 41.7$\pm$0.3 & 38.7$\pm$0.5 & 43.3$\pm$0.0 & \textbf{45.1$\pm$0.0} \\
    \midrule
    DeepUSPS~\cite{(DeepUSPS)nguyen2019deepusps} & - (\textit{unsupervised}) & 37.2$\pm$0.1 & 34.4$\pm$0.2 & 39.8$\pm$0.1 & 42.1$\pm$0.2 \\
    MOVE~\cite{(MOVE)bielski2022move} & - (\textit{unsupervised}) & 41.4$\pm$0.0 & 38.4$\pm$0.0 & 40.6$\pm$0.1 & 42.7$\pm$0.2 \\
    \bottomrule
  \end{tabular}
  \end{adjustbox}
  \end{center}
  \caption{
    \textbf{Quantitative results of WSSS methods on COCO 2014 validation set according to the saliency map generated using different SOD models and SOD datasets.} $\dagger$ indicates using the VGG-16~\cite{(vgg)simonyan2014very} backbone network in DeepLab-V2~\cite{(DLV2)chen2017deeplab}, otherwise using ResNet-101~\cite{(resnet)he2016deep} backbone. Note that the result of a fully-supervised setting is 55.7\%.
  }
  \label{tab:result_coco2014}
\end{table*}

\paragraph{\textbf{Segmentation Networks.}}
We train a baseline segmentation network using generated pseudo labels in a fully-supervised manner. 
We adopt DeepLab-V2~\cite{(DLV2)chen2017deeplab} with ResNet-101~\cite{(resnet)he2016deep} backbone as our baseline network. 
Additionally, we also provide the results on DeepLab-V3+~\cite{(DLV3+)chen2018encoder} with ResNet-101 backbone, SegFormer with MiT-B5 backbone~\cite{(SegFormer)xie2021segformer}, and K-Net~\cite{(K-Net)zhang2021k} with Swin-L~\cite{(swin)liu2021swin} backbone to demonstrate the impact of the segmentation network.
We strictly follow the official training recipe for each network.
The DenseCRF~\cite{(CRF)krahenbuhl2011efficient} post-processing is applied to the segmentation network output only for VOC 2012 dataset.
We report the mean and standard error of the mean intersection over union (mIoU) score over three experiments with different random seeds.

\subsection{Findings}
\label{section_4_3}

\paragraph{Finding 1: The quality of saliency maps has a significant impact on the performance of WSSS methods.}
We evaluate the performance of seven WSSS methods using various kinds of saliency maps generated from four SOD models trained with different SOD datasets. 
The experimental results in Table \ref{tab:result_voc2012} and Table \ref{tab:result_coco2014} indicate that employing more powerful SOD models ($e.g.,$ VST~\cite{(VST)liu2021visual}) or larger-scale SOD datasets ($e.g.,$ DUTS~\cite{(DUTS)wang2017learning}) can boost the performance of WSSS methods. 
Conversely, using weaker SOD models ($e.g.,$ DSS~\cite{(DSS)hou2017deeply}) or smaller-scale SOD datasets ($e.g.,$ MSRA-B~\cite{(MSRA)liu2010learning}) leads to a drop in the performance of WSSS methods. 
\textit{Our finding highlights the critical role of saliency maps in saliency-guided WSSS and the importance of carefully selecting the pre-trained SOD model for rigorous WSSS research.}

\paragraph{Finding 2: Lack of consistency in the saliency maps used in existing WSSS methods.}
Despite the significant impact of saliency maps on WSSS, we found that the saliency map used in each work is not unified.
Inconsistencies exist in both the SOD model and dataset used, with some methods~\cite{(ICD)fan2020learning,(EME)fan2020employing,(SeeNet)hou2018self,(OAA)jiang2019integral,(groupWSSS)li2021group,(AuxSegNet)xu2021leveraging,(NSR)yao2021non} using DSS~\cite{(DSS)hou2017deeply} and some~\cite{(L2G)jiang2022l2g,(DRS)kim2021discriminative,(MCIS)sun2020mining,(EDAM)wu2021embedded,(MOR)zhang2020splitting} using PoolNet~\cite{(PoolNet)liu2019simple}, and some~\cite{(PPC)du2022weakly,(EPS)lee2021railroad,(RCA)zhou2022regional} using PFAN~\cite{(PFAN)zhao2019pyramid} as their SOD model, and with unclear specifications regarding the SOD dataset. 
We cross-validate the performance of WSSS methods using the provided saliency map by each work and found that the performance largely varies depending on the saliency map used (Figure~\ref{fig:cross_validation}).
Even though some methods used the same SOD model, there are non-trivial performance variations (Figure~\ref{fig:sal_comparison}).
\textit{This inconsistency may interfere with meaningful research direction, emphasizing the need for a unified approach in selecting saliency maps for WSSS research.}

\paragraph{\textbf{Finding 3: The conventional CAM can achieve state-of-the-art performance when carefully tuning the threshold $\tau$ and utilizing saliency maps.}}
Previous works considered the conventional CAM~\cite{(CAM)zhou2016learning} to be a baseline method in WSSS, as it appeared to highlight only sparse discriminative object regions, as shown in \figurename~\ref{fig:cam_and_pseudo_labels}.
However, we show that by using a low threshold $\tau$, CAM can capture class-wise information sufficiently.
Although lowering the threshold may produce many false-positives in pseudo labels, these can be filtered out using saliency maps.
According to the threshold, we measure the mIoU score for each WSSS method using the same saliency map and found that the optimal threshold value varies significantly among methods, as shown in \figurename~\ref{fig:threshold_graph}.
Notably, the CAM with a threshold of 0.1 achieves a competitive performance of 71.4\% compared to the state-of-the-art performance of 72.1\%.
Surprisingly, when using K-Net~\cite{(K-Net)zhang2021k} segmentation network, the CAM can yield an unprecedented mIoU score of 79.2\%, surpassing the state-of-the-art, L2G~\cite{(L2G)jiang2022l2g}, which achieves 76.8\% (Table \ref{tab:seg_net_voc12}).
\textit{Our finding emphasizes the need for careful tuning of the threshold in WSSS, an aspect that has received less attention in previous works.}

\begin{table*}[t]
  \begin{center}
  \begin{adjustbox}{max width=\linewidth}
  \begin{tabular}{c|cc|c|c|c}
    \toprule
    Network (Backbone) & SOD Model & SOD Dataset & CAM~\cite{(CAM)zhou2016learning} & EPS~\cite{(EPS)lee2021railroad} & L2G~\cite{(L2G)jiang2022l2g} \\
    \midrule
    \multirow{5}{*}{\shortstack{DeepLab-V2~\cite{(DLV2)chen2017deeplab} (R-101~\cite{(resnet)he2016deep})\\ \footnotesize{(fully-supervised: 77.9\%)}}} & PoolNet~\cite{(PoolNet)liu2019simple} & MSRA-B~\cite{(MSRA)liu2010learning}) & 62.9$\pm$0.6 & 71.4$\pm$0.1 & 70.9$\pm$0.2\\
     & PoolNet~\cite{(PoolNet)liu2019simple} & DUTS~\cite{(DUTS)wang2017learning}) & 68.5$\pm$0.1 & 72.7$\pm$0.2 & 71.7$\pm$0.2 \\
     & VST~\cite{(VST)liu2021visual} & DUTS~\cite{(DUTS)wang2017learning}) & 70.5$\pm$0.2 & 72.4$\pm$0.2 & 71.9$\pm$0.2 \\
     & VST~\cite{(VST)liu2021visual} & COCO-20~\cite{(coco)lin2014microsoft}) & 74.5$\pm$0.2 & 74.3$\pm$0.1 & 73.0$\pm$0.0\\
     & MOVE~\cite{(MOVE)bielski2022move} & \textit{unsupervised} & 69.3$\pm$0.1 & 71.0$\pm$0.1 & 71.0$\pm$0.2 \\
    \midrule
    \multirow{5}{*}{\shortstack{DeepLab-V3+~\cite{(DLV3+)chen2018encoder} (R-101~\cite{(resnet)he2016deep})\\ \footnotesize{(fully-supervised: 78.4\%)}}} & PoolNet~\cite{(PoolNet)liu2019simple} & MSRA-B~\cite{(MSRA)liu2010learning}) & 61.7$\pm$0.2 & 68.5$\pm$0.1 & 69.6$\pm$0.2 \\
     & PoolNet~\cite{(PoolNet)liu2019simple} & DUTS~\cite{(DUTS)wang2017learning}) & 69.4$\pm$0.1 & 70.9$\pm$0.1 & 69.8$\pm$0.1 \\
     & VST~\cite{(VST)liu2021visual} & DUTS~\cite{(DUTS)wang2017learning}) & 69.9$\pm$0.1 & 71.1$\pm$0.1 & 70.3$\pm$0.3 \\
     & VST~\cite{(VST)liu2021visual} & COCO-20~\cite{(coco)lin2014microsoft}) & 74.0$\pm$0.5 & 72.6$\pm$0.1 & 71.6$\pm$0.2 \\
     & MOVE~\cite{(MOVE)bielski2022move} &\textit{unsupervised} & 67.8$\pm$0.4 & 69.8$\pm$0.2 & 68.3$\pm$0.2 \\
    \midrule
    \multirow{5}{*}{\shortstack{SegFormer~\cite{(SegFormer)xie2021segformer} (MiT-B5~\cite{(SegFormer)xie2021segformer})\\ \footnotesize{(fully-supervised: 82.6\%)}}} & PoolNet~\cite{(PoolNet)liu2019simple} & MSRA-B~\cite{(MSRA)liu2010learning}) & 65.4$\pm$0.2 & 71.9$\pm$0.3 & 72.2$\pm$0.2 \\
    & PoolNet~\cite{(PoolNet)liu2019simple} & DUTS~\cite{(DUTS)wang2017learning}) & 72.0$\pm$0.1 & 73.9$\pm$0.3 & 73.5$\pm$0.3 \\
    & VST~\cite{(VST)liu2021visual} & DUTS~\cite{(DUTS)wang2017learning}) & 73.1$\pm$0.3 & 74.0$\pm$0.5 & 74.1$\pm$0.6 \\
    & VST~\cite{(VST)liu2021visual} & COCO-20~\cite{(coco)lin2014microsoft}) & 77.5$\pm$0.1 & 74.9$\pm$0.0 & 75.0$\pm$0.1 \\
    & MOVE~\cite{(MOVE)bielski2022move} & \textit{unsupervised} & 70.7$\pm$0.4 & 72.1$\pm$0.3 & 72.5$\pm$0.2 \\
    \midrule
    \multirow{5}{*}{\shortstack{K-Net~\cite{(K-Net)zhang2021k} (Swin-L~\cite{(swin)liu2021swin})\\ \footnotesize{(fully-supervised: 84.4\%)}}} & PoolNet~\cite{(PoolNet)liu2019simple} & MSRA-B~\cite{(MSRA)liu2010learning}) & 66.7$\pm$0.2 & 73.1$\pm$0.0 & 73.9$\pm$0.0\\
    & PoolNet~\cite{(PoolNet)liu2019simple} & DUTS~\cite{(DUTS)wang2017learning}) & 73.1$\pm$0.6 & 75.0$\pm$0.6 & 74.7$\pm$0.2 \\
     & VST~\cite{(VST)liu2021visual} & DUTS~\cite{(DUTS)wang2017learning}) & 74.2$\pm$0.3 & 75.2$\pm$0.2 & 75.3$\pm$0.5 \\
     & VST~\cite{(VST)liu2021visual} & COCO-20~\cite{(coco)lin2014microsoft}) & \textbf{79.2$\pm$0.1} & 76.8$\pm$0.0 & 76.8$\pm$0.1\\
    & MOVE~\cite{(MOVE)bielski2022move} & \textit{unsupervised} & 72.2$\pm$0.0 & 73.2$\pm$0.4 & 73.8$\pm$0.2 \\
    \bottomrule
  \end{tabular}
  \end{adjustbox}
  \end{center}
  \vspace{-2mm}
  \caption{
    \textbf{Quantitative results of WSSS methods on VOC 2012 validation set} according to the segmentation network and the saliency map generated using different SOD models and SOD datasets. We measure the mean and standard error of the mean intersection over union (mIoU) score over three experiments with different random seeds.
  }
  \label{tab:seg_net_voc12}
  \vspace{4mm}
\end{table*}

\begin{table*}[t]
  \begin{center}
  \begin{adjustbox}{max width=\linewidth}
  \begin{tabular}{c|cc|c|c|c}
    \toprule
    Network (Backbone) & SOD Model & SOD Dataset & CAM~\cite{(CAM)zhou2016learning} & EPS~\cite{(EPS)lee2021railroad} & L2G~\cite{(L2G)jiang2022l2g} \\
    \midrule
    \multirow{4}{*}{\shortstack{DeepLab-V2~\cite{(DLV2)chen2017deeplab} (R-101~\cite{(resnet)he2016deep})\\ \footnotesize{(fully-supervised: 55.7\%)}}} & PoolNet~\cite{(PoolNet)liu2019simple} & MSRA-B~\cite{(MSRA)liu2010learning}) & 31.7$\pm$0.5 & 40.3$\pm$0.1 & 41.8$\pm$0.1\\
    & PoolNet~\cite{(PoolNet)liu2019simple} & DUTS~\cite{(DUTS)wang2017learning}) & 38.7$\pm$0.3 & 43.2$\pm$0.0 & 43.9$\pm$0.1 \\
     & VST~\cite{(VST)liu2021visual} & DUTS~\cite{(DUTS)wang2017learning}) & 41.7$\pm$0.3 & 43.3$\pm$0.0 & 45.1$\pm$0.0\\
     & MOVE~\cite{(MOVE)bielski2022move} & \textit{unsupervised} & 41.4$\pm$0.0 & 40.6$\pm$0.1 & 42.7$\pm$0.2 \\
    \midrule
    \multirow{4}{*}{\shortstack{DeepLab-V3+~\cite{(DLV3+)chen2018encoder} (R-101~\cite{(resnet)he2016deep})\\ \footnotesize{(fully-supervised: 56.6\%)}}} & PoolNet~\cite{(PoolNet)liu2019simple} & MSRA-B~\cite{(MSRA)liu2010learning}) & 30.2$\pm$0.2 & 39.7$\pm$0.0 & 41.8$\pm$0.1 \\
    & PoolNet~\cite{(PoolNet)liu2019simple} & DUTS~\cite{(DUTS)wang2017learning}) & 37.9$\pm$0.0 & 42.7$\pm$0.2 & 43.3$\pm$0.0 \\
     & VST~\cite{(VST)liu2021visual} & DUTS~\cite{(DUTS)wang2017learning}) & 40.4$\pm$0.1 & 42.4$\pm$0.1 & 45.1$\pm$0.1 \\
     & MOVE~\cite{(MOVE)bielski2022move} &\textit{unsupervised} & 40.1$\pm$0.2 & 40.0$\pm$0.0 & 42.2$\pm$0.0 \\
    \midrule
    \multirow{4}{*}{\shortstack{SegFormer~\cite{(SegFormer)xie2021segformer} (MiT-B5~\cite{(SegFormer)xie2021segformer})\\ \footnotesize{(fully-supervised: 65.5\%)}}} & PoolNet~\cite{(PoolNet)liu2019simple} & MSRA-B~\cite{(MSRA)liu2010learning}) & 30.7$\pm$0.1 & 39.5$\pm$0.0 & 42.4$\pm$0.0 \\
    & PoolNet~\cite{(PoolNet)liu2019simple} & DUTS~\cite{(DUTS)wang2017learning}) & 38.8$\pm$0.2 & 43.8$\pm$0.1 & 44.7$\pm$0.1 \\
    & VST~\cite{(VST)liu2021visual} & DUTS~\cite{(DUTS)wang2017learning}) & 41.9$\pm$0.1 & 43.6$\pm$0.1 & 46.0$\pm$0.0 \\
    & MOVE~\cite{(MOVE)bielski2022move} & \textit{unsupervised} & 41.0$\pm$0.0 & 40.8$\pm$0.1 & 43.4$\pm$0.1 \\
    \midrule
    \multirow{4}{*}{\shortstack{K-Net~\cite{(K-Net)zhang2021k} (Swin-L~\cite{(swin)liu2021swin})\\ \footnotesize{(fully-supervised: 67.9\%)}}} & PoolNet~\cite{(PoolNet)liu2019simple} & MSRA-B~\cite{(MSRA)liu2010learning}) & 32.5$\pm$0.5 & 41.7$\pm$0.1 & 43.8$\pm$0.1\\
    & PoolNet~\cite{(PoolNet)liu2019simple} & DUTS~\cite{(DUTS)wang2017learning}) & 40.8$\pm$0.3 &  45.2$\pm$0.1 & 46.4$\pm$0.0  \\
     & VST~\cite{(VST)liu2021visual} & DUTS~\cite{(DUTS)wang2017learning}) & 43.6$\pm$0.2 & 45.3$\pm$0.4 & \textbf{48.4$\pm$0.5} \\
    & MOVE~\cite{(MOVE)bielski2022move} & \textit{unsupervised} & 42.8$\pm$0.5 & 42.5$\pm$0.2 & 44.6$\pm$0.1 \\
    \bottomrule
  \end{tabular}
  \end{adjustbox}
  \end{center}
  \vspace{-2mm}
  \caption{
      \textbf{Quantitative results of WSSS methods on COCO 2014 validation set} according to the segmentation network and the saliency map generated using different SOD models and SOD datasets. We measure the mean and standard error of the mean intersection over union (mIoU) score over three experiments with different random seeds.
  }
  \label{tab:seg_net_coco14}
\end{table*}

\subsection{Discussions}
\label{sec:discussion}

In this part, we further investigate the impact of saliency maps on WSSS through additional studies.
\begin{figure*}[t]
    \centering
    \includegraphics[width=\linewidth]{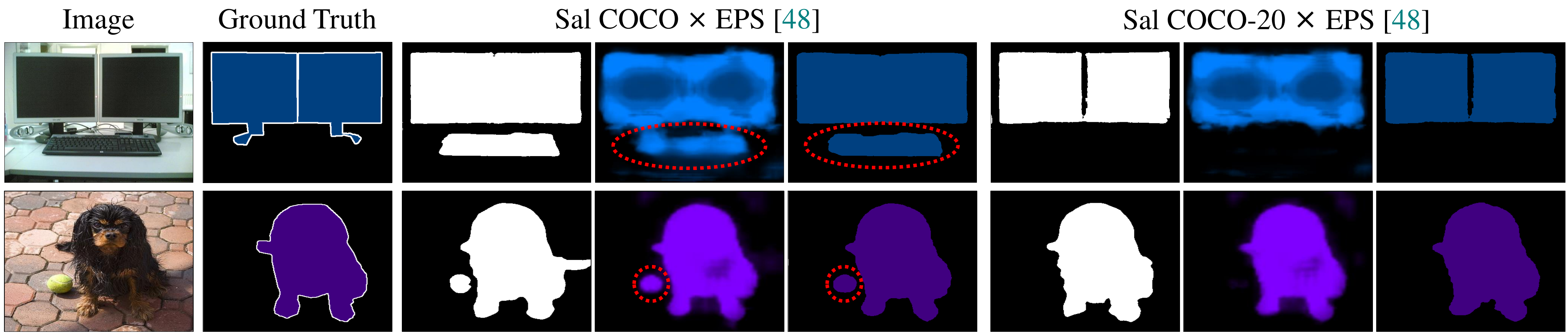}
    \caption{
        \textbf{Qualitative samples for domain gap between the SOD dataset and the target WSSS dataset.} Given the saliency map, we visualize the activation map from EPS~\cite{(EPS)lee2021railroad} and the pseudo label. The SOD model trained with COCO dataset indicates the \textit{keyboard} and \textit{tennis ball} as salient, but the target WSSS dataset ($i.e.,$ VOC 2012) regards those objects as background, incurring false-positives in the pseudo label. 
        When using COCO-20 dataset where only classes in VOC 2012 are treated as salient, the domain gap is resolved, generating the adequate pseudo label for VOC 2012.
    }
    \label{fig:domain_gap}
\end{figure*}

\paragraph{Can we improve the performance of the WSSS methods in general by using more accurate saliency maps?}
We question whether using more accurate saliency maps further improves the performance of WSSS methods.
To make further higher-quality saliency maps, we train the SOD model using a much larger dataset, COCO 2017 dataset~\cite{(coco)lin2014microsoft} which contains 115K images.
We transform the class-wise object masks in COCO into class-agnostic salient object masks. 
As shown in Table~\ref{tab:result_voc2012}, when using the COCO dataset as the SOD dataset, the performance of WSSS methods on VOC 2012 benchmark significantly improves, surpassing state-of-the-art results and achieving 74.8\% of EDAM~\cite{(EDAM)wu2021embedded} compared to 77.9\% of the fully-supervised setting.
This highlights that leveraging significantly improved saliency maps can lead to remarkable performance improvement in the WSSS.
We note that the purpose of using COCO dataset is not to compete with other methods in academic fields but to show the potential for improvement of WSSS in practical scenarios.

\paragraph{\textbf{Further boosting the performance of WSSS methods using advanced segmentation networks.}}
We analyze the effect of the segmentation network on the performance of WSSS methods using three additional segmentation networks, DeepLab-V3+~\cite{(DLV3+)chen2018encoder} with ResNet-101 backbone, SegFormer~\cite{(SegFormer)xie2021segformer} with MiT-B5~\cite{(SegFormer)xie2021segformer} backbone, and K-Net~\cite{(K-Net)zhang2021k} with Swin-L~\cite{(swin)liu2021swin} backbone.
We train the networks following their official training recipe using MMSegmentation~\cite{mmseg2020}.
For the analysis, we adopt three WSSS methods, CAM~\cite{(CAM)zhou2016learning}, EPS~\cite{(EPS)lee2021railroad}, and L2G~\cite{(L2G)jiang2022l2g}.
As the results in Table~\ref{tab:seg_net_voc12} and Table~\ref{tab:seg_net_coco14}, the segmentation network has a significant impact on the performance of WSSS methods.
Namely, employing the better segmentation network ($i.e.,$ K-Net) leads to a substantial performance improvement, even surpassing the result of the fully-supervised DeepLab-V2 with ResNet-101 backbone.
This result demonstrates that using a better segmentation network can further improve the performance of WSSS methods.

\paragraph{What impact does the domain gap between the SOD dataset and the target WSSS dataset has?}
We notice that salient objects are treated differently across WSSS datasets.
For instance, when the saliency map identifies \textit{keyboard} as salient, the VOC dataset treats it as background, while the COCO dataset treats it as a valid class, as shown in Figure~\ref{fig:domain_gap}.
This domain gap can lead to artifacts in saliency-guided WSSS methods since they heavily rely on the knowledge in the saliency map. 
To quantitatively analyze the impact of the domain gap in WSSS, we create the COCO-20 dataset, where the object masks in the COCO 2017 dataset are converted to salient masks only for the twenty VOC classes. 
The experimental results on VOC 2012 in Table~\ref{tab:result_voc2012} demonstrate that the performance of the WSSS methods is higher when using the COCO-20 dataset than when using the COCO dataset. 
This highlights the domain gap between the SOD dataset and the target WSSS dataset is one of the factors to be considered in saliency-guided WSSS methods.

\begin{figure*}[t]
    \centering
    \includegraphics[width=\linewidth]{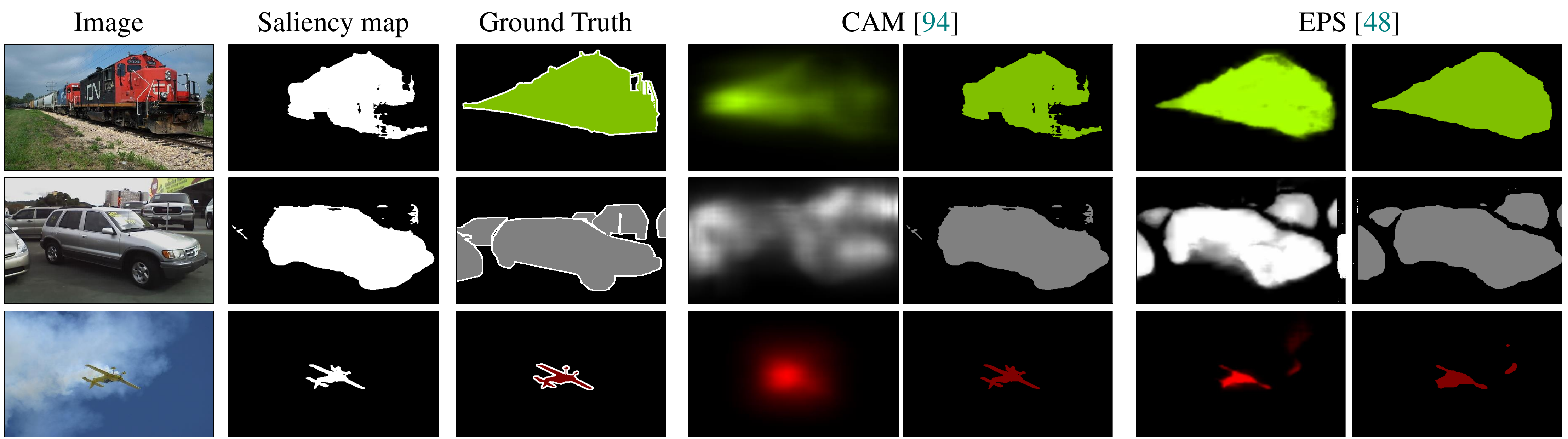}
    \caption{\textbf{Qualitative samples of robustness to the saliency map.} Some methods ($e.g.,$ CAM~\cite{(CAM)zhou2016learning}, DRS~\cite{(DRS)kim2021discriminative}) that utilize the saliency map when generating pseudo labels are more vulnerable to errors in the saliency map, whereas some methods ($e.g.,$ EPS~\cite{(EPS)lee2021railroad} and L2G~\cite{(L2G)jiang2022l2g}) that employ the saliency when training the classifier can generalize errors in the saliency map (1st and 2nd rows). However, when the saliency map is good enough, the pseudo label from CAM can be superior to that from EPS.
    }
    \label{fig:generalization}
\end{figure*}

\paragraph{How to exploit the saliency map?}
There exist two types of methods for exploiting the saliency map: \textit{saliency-pseudo-labeling methods} that use it only when generating pseudo labels and \textit{saliency-supervised methods} that incorporate it as a supervision source for training the classifier without using it when generating pseudo labels.
As the results in Table~\ref{tab:result_voc2012} and Table~\ref{tab:result_coco2014}, we observe that the saliency-pseudo-labeling methods ($e.g.,$ CAM~\cite{(CAM)zhou2016learning}, OAA~\cite{(OAA)jiang2019integral}, MCIS~\cite{(MCIS)sun2020mining}, DRS~\cite{(DRS)kim2021discriminative}, and EDAM~\cite{(EDAM)wu2021embedded}) heavily rely on the saliency maps. In contrast, the saliency-supervised methods ($e.g.,$ EPS~\cite{(EPS)lee2021railroad} and L2G~\cite{(L2G)jiang2022l2g}) are more robust to the saliency maps since the saliency-guided classifier can generalize and correct some errors in the saliency map (1st and 2nd rows in Figure~\ref{fig:generalization}).
Furthermore, the saliency-supervised methods can generate activation maps that more accurately align with the object region than the saliency-pseudo-labeling methods.
However, the saliency-supervised methods may fail to generalize or represent detailed information (3rd row in Figure~\ref{fig:generalization}).

\subsection{Recommendations}

To encourage more meaningful and rigorous research for WSSS, we recommend the following things.

\paragraph{\textbf{Recommendation 1: Use the same saliency maps across all compared models and specify the sources.}}
As discussed in the Findings section, we observed the significant impact of saliency maps on WSSS and found inconsistency in saliency maps among different WSSS works.
To this end, we introduce a new framework called \textbf{\texttt{WSSS-BED}} that enables researchers to conduct experiments under unified conditions.
By submitting their class activation maps to \texttt{WSSS-BED}, it generates pseudo segmentation labels using unified saliency maps.
An early version of \texttt{WSSS-BED} is publicly available at \url{https://github.com/clovaai/WSSS-BED}, which includes various saliency maps used in Table~\ref{tab:result_voc2012} and Table~\ref{tab:result_coco2014} and activation maps of seven WSSS methods.
\texttt{WSSS-BED} will be regularly updated upon pull request from fellow researchers.
We encourage researchers to use \texttt{WSSS-BED} and specify the source of the saliency map to ensure the consistency and reproducibility of their experiments.

\paragraph{\textbf{Recommendation 2: Thoroughly study the effect of the threshold and saliency maps.}}
We provided new insights that both threshold and saliency maps are non-trivial factors in WSSS, as shown in \figurename~\ref{fig:cross_validation} and \figurename~\ref{fig:threshold_graph}.
For the rigorous development of WSSS, we encourage fellow researchers to study the effect of the threshold and saliency maps on their method thoroughly.
The ideal threshold and saliency maps for each method may be different, and \texttt{WSSS-BED} facilitates this endeavor.

\paragraph{\textbf{Recommendation 3: Utilize unsupervised salient object detection (SOD) models.}}
Although pre-trained SOD models are widely available and commonly used in various research fields, such as weakly-supervised learning~\cite{(AE-PSL)wei2017object}, data augmentation~\cite{kim2020puzzle}, and continual learning~\cite{cha2021ssul}, using fully-supervised SOD models in weakly-supervised learning may deviate from the intended setting since it requires fully-annotated class-agnostic object masks for pre-training.
To promote a more desirable weakly-supervised setting, we recommend using saliency maps from unsupervised SOD models, which do not require fully-annotated object masks.
\texttt{WSSS-BED} provides saliency maps from two unsupervised SOD models, DeepUSPS~\cite{(DeepUSPS)nguyen2019deepusps} and MOVE~\cite{(MOVE)bielski2022move}.
As shown in Table~\ref{tab:result_voc2012} and Table~\ref{tab:result_coco2014}, leveraging unsupervised SOD models can achieve reasonable performance while maintaining a fully weakly-supervised setting, which might be an alternative to compare with saliency-free approaches.

\section{Related Work}

\subsection{Salient Object Detection}
Salient object detection (SOD) aims to identify the most visually distinctive objects in an image and commonly serves as a basic technique for various computer vision applications, such as visual tracking~\cite{hong2015online} and image retrieval~\cite{gao20123}.
SOD approaches can be categorized as supervised~\cite{(DSS)hou2017deeply,(PoolNet)liu2019simple,(VST)liu2021visual,(PFAN)zhao2019pyramid} or unsupervised~\cite{(MOVE)bielski2022move,(DeepUSPS)nguyen2019deepusps} methods depending on whether they require fully-annotated salient object masks; we provide the detail information about it in our appendix.

\subsection{Weakly-Supervised Semantic Segmentation}
Weakly-supervised semantic segmentation (WSSS) approaches can be divided into two groups based on whether they use saliency maps as off-the-shelf knowledge or not.
We briefly summarize each group in this section and deeply discuss each method in our appendix. \\

\noindent \textbf{Saliency-guided approaches.} 
Saliency-guided approaches often address the problem of sparse activation maps of CAM~\cite{(CAM)zhou2016learning} and proposed methods to cover the object region entirely using erasing~\cite{(SeeNet)hou2018self,(FickleNet)lee2019ficklenet,(AE-PSL)wei2017object}, expanding~\cite{(DSRG)huang2018weakly,(DRS)kim2021discriminative,(MDC)wei2018revisiting}, graph neural network~\cite{(groupWSSS)li2021group,(NSR)yao2021non}, or self-attention~\cite{(EDAM)wu2021embedded,(SGAN)yao2020saliency}. \\

\noindent \textbf{Saliency-free approaches.} 
Without the off-the-shelf knowledge of the saliency map, saliency-free approaches have proposed novel methods based on affinity learning~\cite{(affinitynet)ahn2018learning,(SANCE)li2022towards}, contrastive learning~\cite{(PPC)du2022weakly,(RCA)zhou2022regional}, graph neural network~\cite{pan2021weakly}, self-attention~\cite{(SEAM)Wang2020SEAM}, causal inference~\cite{(CONTA)dong2020conta}, anti-adversarial attack~\cite{(AdvCAM)Lee2021AdvCAM}, conditional random field~\cite{(PMM)li2021pseudo,(RRM)zhang2020reliability}.
Recently, AMN~\cite{(AMN)lee2022threshold} has demonstrated the importance of the threshold in saliency-free WSSS.
However, our observations and results are completely different from theirs since our analysis is based on using saliency maps; the optimal threshold of CAM in our setting is lower than theirs, which allows us to further improve the performance of CAM.
\section{Conclusion}

This paper provided a new perspective on the role of saliency maps in weakly supervised semantic segmentation (WSSS) based on empirical observations.
We highlighted that the quality of saliency maps is a crucial factor in saliency-guided WSSS and revealed that the saliency maps used in existing methods are not unified.
Additionally, we provided some discussion points for future research direction, such as the choice of the threshold in WSSS, the domain gap between the saliency map and target dataset, and how to use the saliency map.
To facilitate more meaningful and rigorous research for saliency-guided WSSS, we introduced a new framework called \texttt{WSSS-BED}.
\texttt{WSSS-BED} enables conducting research under unified conditions and contains various saliency maps and activation maps for seven WSSS methods, which will be continuously updated with available models upon pull request from researchers.
We hope our new empirical findings enable more rigorous research and promote meaningful advances in saliency-guided WSSS.

\nocite{(domain-bed)gulrajani2021in}
\nocite{he2019rethinking}
\nocite{choe2020evaluating}
\nocite{kim2023devil}

{\small
\bibliographystyle{ieeenat_fullname}
\bibliography{ms}
}

\clearpage
\appendix
\renewcommand{\thesection}{\Alph{section}}
\section*{Appendix}

\section{Additional Analysis}

\paragraph{\textbf{Numerical results of performance variation according to the saliency map.}}
In Figure \ref{fig:cross_validation}, we revealed the saliency map used in each method is not unified, and the effect of the saliency map on each method is highly significant.
Here, Table~\ref{tab:cross_validate_voc2012} provides all numerical results of Figure \ref{fig:cross_validation}.

\paragraph{\textbf{Numerical results for the effect of the threshold.}}
In Figure \ref{fig:threshold_graph}, we provided the impact of the threshold on WSSS methods.
Here, Table~\ref{tab:threshold} provides the optimal threshold and numerical result of each method. 

\section{Related Works}
\subsection{Weakly-Supervised Semantic Segmentation}
Weakly-supervised semantic segmentation (WSSS) approaches can be divided into two groups based on whether they use saliency maps as off-the-shelf knowledge or not.

\begin{table*}[t]
  \centering
  \begin{adjustbox}{max width=\linewidth}
  \begin{tabular}{c|c|c|c|c|c|c|c}
    \toprule
    Saliency map & CAM~\cite{(CAM)zhou2016learning}& OAA~\cite{(OAA)jiang2019integral} & MCIS~\cite{(MCIS)sun2020mining} & DRS~\cite{(DRS)kim2021discriminative} & EDAM~\cite{(EDAM)wu2021embedded} & EPS~\cite{(EPS)lee2021railroad} & L2G~\cite{(L2G)jiang2022l2g} \\
    \midrule
    \textit{Paper} & - & 65.2 & 66.2 & 70.4 & 70.9 & 70.9 & 72.1 \\
    \midrule
    sal OAA (DSS~\cite{(DSS)hou2017deeply}) & 69.3$\pm$0.0 & 68.7$\pm$0.3 & 68.0$\pm$0.2 & 68.6$\pm$0.1 & 69.5$\pm$0.1 & 70.3$\pm$0.1 & 71.1$\pm$0.1 \\
    sal MCIS (DSS~\cite{(DSS)hou2017deeply}) & 69.3$\pm$0.0 & 68.7$\pm$0.3 & 68.0$\pm$0.2 & 68.6$\pm$0.1 & 69.5$\pm$0.1 & 70.3$\pm$0.1 & 71.1$\pm$0.1 \\
    sal DRS (PoolNet~\cite{(PoolNet)liu2019simple}) & 71.2$\pm$0.1 & 70.7$\pm$0.2 & 70.3$\pm$0.2 & 70.7$\pm$0.2 & 71.7$\pm$0.2 & 71.9$\pm$0.1 & 71.5$\pm$0.3 \\
    sal EDAM (PoolNet~\cite{(PoolNet)liu2019simple}) & 70.4$\pm$0.1 & 69.6$\pm$0.2 & 69.0$\pm$0.1 & 70.0$\pm$0.1 & 70.0$\pm$0.2 & 71.7$\pm$0.1 & 71.6$\pm$0.2 \\
    sal EPS (PFAN~\cite{(PFAN)zhao2019pyramid}) & 65.7$\pm$0.3 & 64.5$\pm$0.2 & 64.5$\pm$0.3 & 65.4$\pm$0.3 & 66.0$\pm$0.3 & 71.1$\pm$0.1 & 70.6$\pm$0.1 \\
    sal L2G (PoolNet~\cite{(PoolNet)liu2019simple}) & 68.5$\pm$0.1 & 68.7$\pm$0.3 & 67.1$\pm$0.1 & 68.2$\pm$0.0 & 68.3$\pm$0.1 & 72.0$\pm$0.1 & 72.2$\pm$0.2 \\
    \bottomrule
  \end{tabular}
  \end{adjustbox}
  \caption{\textbf{Performance variation according to the saliency map.} We discover that the saliency map used in each method is not unified, and the impact of the saliency map on each method is highly significant. `\textit{Paper}' in the second row means the score is from their paper. `Sal \texttt{METHOD} (\texttt{SOD-MODEL})' indicates saliency maps used in the WSSS \texttt{METHOD} and they used \texttt{SOD-MODEL} when generating the saliency maps. The scores are measured on VOC 2012 validation set.}
  \label{tab:cross_validate_voc2012}
\end{table*}

\begin{table}[t]
  \centering
  \begin{adjustbox}{max width=\linewidth}
  \begin{tabular}{c|c|c}
    \toprule
    Method & Threshold & mIoU \\
    \midrule
    CAM~\cite{(CAM)zhou2016learning} & 0.1 & 71.2 \\
    OAA~\cite{(OAA)jiang2019integral} \textsubscript{ICCV'19} & 0.3 & 70.7 \\
    MCIS~\cite{(MCIS)sun2020mining} \textsubscript{ECCV'20} & 0.5 & 70.3 \\
    DRS~\cite{(DRS)kim2021discriminative} \textsubscript{AAAI'21} & 0.5 & 70.7 \\
    EDAM~\cite{(EDAM)wu2021embedded} \textsubscript{CVPR'21} & 0.5 & 71.7 \\
    EPS~\cite{(EPS)lee2021railroad} \textsubscript{CVPR'21} & 0.2 & 71.9 \\
    L2G~\cite{(L2G)jiang2022l2g} \textsubscript{CVPR'22} & 0.3 & 71.5 \\
    \bottomrule
  \end{tabular}
  \end{adjustbox}
  \caption{
    \textbf{The performance comparisons of WSSS methods according to the threshold $\tau$} on \textit{VOC2012} validation set using the same saliency maps. The optimal threshold varies from each work, and the conventional CAM with a low threshold shows a highly competitive performance compared to the state-of-the-art method.
  }
  \label{tab:threshold}
\end{table}

\paragraph{Saliency-guided approaches.}
Saliency-guided methods often address the problem of sparse activation maps of CAM~\cite{(CAM)zhou2016learning} and propose methods to cover the object region entirely.
AE-PSL~\cite{(AE-PSL)wei2017object} {\scriptsize (CVPR'17)} proposed an adversarial erasing approach for expanding object regions.
DSRG~\cite{(DSRG)huang2018weakly} {\scriptsize (CVPR'18)} introduced the seeded region growing module to densely cover the object region.
MDC~\cite{(MDC)wei2018revisiting} {\scriptsize (CVPR'18)} presented multiple dilated convolutional blocks to produce dense object localization maps.
MCOF~\cite{(MCOF)wang2018weakly} {\scriptsize (CVPR'18)} proposed an iterative bottom-up and top-down framework to expand object regions.
SeeNet\cite{(SeeNet)hou2018self} {\scriptsize (NeurIPS'18)} introduced a self-erasing network to prevent attention from spreading to undesired background regions.
FickleNet~\cite{(FickleNet)lee2019ficklenet} {\scriptsize (CVPR'19)} proposed a dropout method to generate multiple localization maps from a single image.
SGAN~\cite{(SGAN)yao2020saliency} {\scriptsize (ACCESS)} introduced a saliency-guided self-attention network to capture rich and extensive contextual information.
OAA~\cite{(OAA)jiang2019integral} {\scriptsize (ICCV'19)} presented an online attention accumulation strategy to accumulate attention on different object parts.
MCIS~\cite{(MCIS)sun2020mining} {\scriptsize (ECCV'20)} introduced a co-attention classification network to discover integral object regions by addressing cross-image semantics.
MOR~\cite{(MOR)zhang2020splitting} {\scriptsize (ECCV'20)} proposed a splitting vs. merging optimization strategy to mine out regions of different spatial patterns and the common regions of the different maps.
EM~\cite{(EME)fan2020employing} {\scriptsize (ECCV'20)} introduced an approach to alleviate the inaccurate seed problem by leveraging the robustness of the segmentation model to learn from multiple seeds.
GroupWSSS~\cite{(groupWSSS)li2021group} {\scriptsize (AAAI'21)} utilized a graph neural network that can operate on a group of images and explore their semantic relations for representation learning.
DRS~\cite{(DRS)kim2021discriminative} {\scriptsize (AAAI'21)} introduced a discriminative region suppression module to suppress the attention on discriminative regions and spread it to adjacent non-discriminative regions.
NSR~\cite{(NSR)yao2021non} {\scriptsize (CVPR'21)} proposed a graph-based global reasoning unit to strengthen the classification network to capture global relations among disjoint and distant regions.
EPS~\cite{(EPS)lee2021railroad} {\scriptsize (CVPR'21)} presented a framework that learns from pixel-level feedback from image-level labels and saliency maps.
EDAM~\cite{(EDAM)wu2021embedded} {\scriptsize (CVPR'21)} proposed an embedded discriminative attention mechanism to explore more comprehensive class-specific activation maps.
AuxSegNet~\cite{(AuxSegNet)xu2021leveraging} {\scriptsize (ICCV'21)} proposed a framework to leverage saliency detection and multi-label image classification as auxiliary tasks.
PPC~\cite{(PPC)du2022weakly} {\scriptsize (CVPR'22)} proposed weakly-supervised pixel-to-prototype contrast that can provide pixel-level supervisory signals.
RCA~\cite{(RCA)zhou2022regional} {\scriptsize (CVPR'22)} presented regional semantic contrast and aggregation to explore rich semantic contexts synergistically among abundant weakly-labeled training data for network learning and inference.
L2G~\cite{(L2G)jiang2022l2g} {\scriptsize (CVPR'22)} proposed a local-to-global attention transfer method to attain object attention.

We note that the adopted salient object detection (SOD) model is different for each WSSS method as shown in Table~\ref{tab:methods_and_SOD_models} and the quality of saliency maps may vary even if the same SOD model is used as we discussed before.

\paragraph{Saliency-free approaches.}
Without using the off-the-shelf knowledge of the saliency map, saliency-free approaches have proposed novel methods.
AffinityNet~\cite{(affinitynet)ahn2018learning} {\scriptsize (CVPR'18)} proposed a network that predicts semantic affinity between a pair of adjacent image coordinates.
SSDD~\cite{(SSDD)shimoda2019self} {\scriptsize (ICCV'19)} presented a self-supervised difference detection module that estimates noise from the results of the mapping functions by predicting the difference between the segmentation masks before and after the mapping.
RRM~\cite{(RRM)zhang2020reliability} {\scriptsize (AAAI'20)} proposed a one-step approach through mining reliably yet tiny regions and using them as ground-truth labels directly for segmentation model training.
SEAM~\cite{(SEAM)Wang2020SEAM} {\scriptsize (CVPR'20)} introduced a self-supervised equivariant attention mechanism to discover additional supervision.
SingleStage~\cite{(SSS)araslanov2020single} {\scriptsize (CVPR'20)} presented a segmentation-based network model and a self-supervised training scheme to train for semantic masks in a single stage.
ICD~\cite{(ICD)fan2020learning} {\scriptsize (CVPR'20)} proposed an efficient end-to-end intra-class discriminator framework that learns intra-class boundaries to help separate the foreground and the background within each image-level class.
SCE~\cite{(SCE)chang2020weakly} {\scriptsize (CVPR'20)} introduced a self-supervised task by exploiting the sub-category information.
BES~\cite{(BES)Chen2020BES} {\scriptsize (ECCV'20)} proposed an approach to explicitly explore object boundaries to refine coarse localization maps for training images.
CONTA~\cite{(CONTA)dong2020conta} {\scriptsize (NeurIPS'20)} presented a structural causal model to analyze the causalities among images, contexts, and class labels.
AdvCAM~\cite{(AdvCAM)Lee2021AdvCAM} {\scriptsize (CVPR'21)} proposed an attribution map of an image that is manipulated to increase the classification score, allowing it to identify more regions of an object.
CGNet~\cite{(CGNet)kweon2021unlocking} {\scriptsize (ICCV'21)} introduced a class-specific adversarial erasing-based framework that fully exploits the potential of a pre-trained classifier.
PMM~\cite{(PMM)li2021pseudo} {\scriptsize (ICCV'21)} introduced proportional pseudo-mask generation with co-efficient of variation smoothing to expand the activation area of CAM and compute the importance of each location for each class independently.
RIB~\cite{(RIB)lee2021reducing} {\scriptsize (NeurIPS'21)} proposed a method to reduce the information bottleneck by removing the last activation function and introduced a pooling method that encourages the transmission of information from non-discriminative regions to the classification.
URN~\cite{li2(URN)022uncertainty} {\scriptsize (AAAI'22)} mitigated the noise in segmentation optimization with the uncertainty from scaling.
AMR~\cite{(AMR)qin2022activation} {\scriptsize (AAAI'22)} proposed an activation modulation and recalibration scheme that leverages a spotlight branch and a compensation branch to obtain weighted CAMs that can provide recalibration supervision and task-specific concepts.
ReCAM~\cite{(ReCAM)Chen2022ClassRM} {\scriptsize (CVPR'22)} proposed a method that plugs softmax cross-entropy loss into the binary cross-entropy based method to reactivate the model.
AFA~\cite{(AFA)ru2022learning} {\scriptsize (CVPR'22)} introduced an affinity-from-attention module to learn semantic affinity from the multi-head self-attention in transformers.
MCTformer~\cite{(MCTformer)xu2022multi} {\scriptsize (CVPR'22)} proposed a multi-class token transformer that uses multiple class tokens to learn interactions between the class tokens and the patch tokens.
SIPE~\cite{(SIPE)chen2022self} {\scriptsize (CVPR'22)} presented a self-supervised image-specific prototype exploration to capture complete activation regions to empower a self-correction ability of prototype exploration.
W-OoD~\cite{(W-OoD)lee2022weakly} {\scriptsize (CVPR'22)} utilized out-of-distribution data or images devoid of foreground object classes to distinguish foreground from the background.
AMN~\cite{(AMN)lee2022threshold} {\scriptsize (CVPR'22)} provided empirical evidence that the actual performance bottleneck of WSSS is a global thresholding scheme applied after CAM and proposed an activation manipulation network with a per-pixel classification loss and a label conditioning module.

\begin{table}[t]
  \centering
  \begin{adjustbox}{max width=\linewidth}
  \begin{tabular}{c|c|c}
    \toprule
    WSSS method & SOD model & mIoU (\%) \\
    \midrule
    AE-PSL~\cite{(AE-PSL)wei2017object} \textsubscript{CVPR'17} & DRFI~\cite{(DRFI)jiang2013salient} & 55.0 \\
    MCOF~\cite{(MCOF)wang2018weakly} \textsubscript{CVPR'18} & DRFI~\cite{(DRFI)jiang2013salient} & 60.3 \\
    MDC~\cite{(MDC)wei2018revisiting} \textsubscript{CVPR'18} & SENet~\cite{(SENet)xiao2017self} & 60.4 \\
    DSRG~\cite{(DSRG)huang2018weakly} \textsubscript{CVPR'18} & DRFI~\cite{(DRFI)jiang2013salient} & 61.4 \\
    SeeNet~\cite{(SeeNet)hou2018self} \textsubscript{NIPS'18} & DSS~\cite{(DSS)hou2017deeply}  & 63.1 \\
    FickleNet~\cite{(FickleNet)lee2019ficklenet} \textsubscript{CVPR'19} & DRFI~\cite{(DRFI)jiang2013salient} & 64.9 \\
    OAA~\cite{(OAA)jiang2019integral} \textsubscript{ICCV'19} & DSS~\cite{(DSS)hou2017deeply}  & 65.2 \\
    SGAN~\cite{(SGAN)yao2020saliency} \textsubscript{ACCESS} & S-Net~\cite{(S-Net)xiao2018deep} & 67.1 \\
    \midrule
    MCIS~\cite{(MCIS)sun2020mining} \textsubscript{ECCV'20} & DSS~\cite{(DSS)hou2017deeply} & 66.2 \\
    MOR~\cite{(MOR)zhang2020splitting} \textsubscript{ECCV'20} & PoolNet~\cite{(PoolNet)liu2019simple} & 66.6 \\
    EM~\cite{(EME)fan2020employing} \textsubscript{ECCV'20} & DSS~\cite{(DSS)hou2017deeply}  & 67.2 \\
    ICD~\cite{(ICD)fan2020learning} \textsubscript{CVPR'20} & DSS~\cite{(DSS)hou2017deeply}  & 67.8 \\
    \midrule
    GroupWSSS~\cite{(groupWSSS)li2021group} \textsubscript{AAAI'21} & DSS~\cite{(DSS)hou2017deeply}  & 68.2 \\
    AuxSegNet~\cite{(AuxSegNet)xu2021leveraging} \textsubscript{ICCV'21} & DSS~\cite{(DSS)hou2017deeply}  & 69.0 \\
    NSR~\cite{(NSR)yao2021non} \textsubscript{CVPR'21} & DSS~\cite{(DSS)hou2017deeply}  & 70.4 \\
    DRS~\cite{(DRS)kim2021discriminative} \textsubscript{AAAI'21} & PoolNet~\cite{(PoolNet)liu2019simple} & 70.4 \\
    EDAM~\cite{(EDAM)wu2021embedded} \textsubscript{CVPR'21} & PoolNet~\cite{(PoolNet)liu2019simple} & 70.9 \\
    EPS~\cite{(EPS)lee2021railroad} \textsubscript{CVPR'21}  & PFAN~\cite{(PFAN)zhao2019pyramid} & 70.9 \\
    \midrule
    L2G~\cite{(L2G)jiang2022l2g} \textsubscript{CVPR'22}  & PoolNet~\cite{(PoolNet)liu2019simple} & 72.1 \\
    RCA~\cite{(RCA)zhou2022regional} \textsubscript{CVPR'22}  & PFAN~\cite{(PFAN)zhao2019pyramid} & 72.2 \\
    PPC~\cite{(PPC)du2022weakly} \textsubscript{CVPR'22}  & PFAN~\cite{(PFAN)zhao2019pyramid} & 72.6 \\
    \bottomrule
  \end{tabular}
  \end{adjustbox}
  \caption{
    The salient object detection (SOD) models used in existing saliency-guided WSSS methods and their official performance on Pascal VOC 2012 validation set.
    }
  \label{tab:methods_and_SOD_models}
\end{table}

\subsection{Salient Object Detection}
Salient object detection (SOD) aims to identify the most visually distinctive objects in an image.
SOD approaches can be categorized as supervised~\cite{(DSS)hou2017deeply,(PoolNet)liu2019simple,(VST)liu2021visual,(PFAN)zhao2019pyramid} or unsupervised~\cite{(MOVE)bielski2022move,(DeepUSPS)nguyen2019deepusps} methods depending on whether they require fully-annotated salient object masks.

\paragraph{\textbf{Supervised methods.}}
Given a dataset consisting of pairs of images and ground-truth salient object masks, the SOD model is trained in a fully-supervised manner.
DSS~\cite{(DSS)hou2017deeply} proposed a multi-level feature fusion strategy with U-Net~\cite{(unet)ronneberger2015u} network architecture.
PFAN~\cite{(PFAN)zhao2019pyramid} leverages channel attention and spatial attention in the multi-level feature pyramid network.
PoolNet~\cite{(PoolNet)liu2019simple} also adopts the U-Net architecture with the proposed feature aggregation module and global guidance module.
Recently, by adopting the ViT~\cite{(vit)dosovitskiy2020image} architecture, VST~\cite{(VST)liu2021visual} designed the multi-level token fusion and token upsampling methods.

\paragraph{\textbf{Unsupervised methods.}}
The unsupervised method does not require ground-truth salient object masks.
DeepUSPS~\cite{(DeepUSPS)nguyen2019deepusps} proposed a method that iteratively refines noisy pseudo saliency maps from different handcrafted methods.
MOVE~\cite{(MOVE)bielski2022move} exploited the fact that foreground objects can be shifted locally relative to their initial position and result in realistic new images.

\end{document}